\documentclass{sig-alternate}

\usepackage{makeidx}

\newif\iftwocol\twocolfalse             
\newif\ifshort\shortfalse               
\newif\iffinal\finaltrue                

\newif\iflonger\longerfalse    

\begin{document}
\conferenceinfo{GECCO'05,} {June 25--29, 2005, Washington, DC, USA.}

\CopyrightYear{2005}

\crdata{1-59593-010-8/05/0006}

\title{Fitness Uniform Deletion:\\A Simple Way to Preserve Diversity}

\iffinal
\numberofauthors{2}
\author{
\alignauthor Shane Legg\\
       \affaddr{IDSIA, Galleria 2, 6928 Manno-Lugano}\\
       \affaddr{Switzerland}\\
       \email{shane@idsia.ch}
\alignauthor Marcus Hutter\\
       \affaddr{IDSIA, Galleria 2, 6928 Manno-Lugano}\\
       \affaddr{Switzerland}\\
       \email{marcus@idsia.ch}
}

\else
\numberofauthors{1}
\author{
\alignauthor Track category: Genetic Algorithms\\
       \affaddr{\mbox{}}\\
       \affaddr{Keywords: Fitness Landscapes, Parameter Tuning, Selection}\\
       \affaddr{\mbox{}}\\
       \affaddr{\mbox{}}\\
       \affaddr{\mbox{}}\\
       \email{\mbox{}}
}
\fi

\maketitle

\begin{abstract}
A commonly experienced problem with population based optimisation
methods is the gradual decline in population diversity that tends
to occur over time.  This can slow a system's progress or even
halt it completely if the population converges on a local optimum
from which it cannot escape.  In this paper we present the Fitness
Uniform Deletion Scheme (FUDS), a simple but somewhat
unconventional approach to this problem.  Under FUDS the deletion
operation is modified to only delete those individuals which are
``common'' in the sense that there exist many other individuals of
similar fitness in the population.  This makes it impossible for
the population to collapse to a collection of highly related
individuals with similar fitness. Our experimental results on a
range of optimisation problems confirm this, in particular for
deceptive optimisation problems the performance is significantly
more robust to variation in the selection intensity.
\end{abstract}

\category{I.2.M}{Artificial Intelligence}{Miscellaneous}
\terms{Algorithms, Experimentation}
\keywords{Genetic algorithms, population diversity}


\section{Introduction}\label{sec:intro}

Population based optimisation methods often suffer from diversity
problems, especially after many generations.  In the worst case the
whole population can become confined to a local optimum from which it
cannot escape, leaving the optimiser stuck.  In situations where the
population size must be kept small due to resource constraints, or the
solution space is deceptive with many large local optima, the loss of
population diversity becomes especially significant.

Numerous techniques have been devised to help preserve population
diversity.  Significant contributions in this direction are fitness
sharing \cite{Goldberg:87}, crowding \cite{DeJong:75}, local mating
\cite{Collins:91} and niching \cite{Mahfoud:95}.  Many of these
methods are based on some kind of distance measure on the solution
space which is used to ensure that the members of the population do
not become too similar to each other.  Unfortunately, for many
interesting optimisation problems, such as evolving neural networks or
genetic programming, just trying to establish whether two individuals
are effectively the same can be very difficult as totally different
neural networks or programs (genotype) can have identical behaviour
(phenotype).  In the most general case when the solution space is
Turing complete this comparison is impossible even in theory.

\vspace{0.05em}

While most methods try to measure diversity directly in genotype
space, an alternative approach is to measure diversity in phenotype
space\iflonger, for example by using the fitness of individuals as a
rough guide to their similarity.\else. \fi An example of this approach
is the Fitness Uniform Selection Scheme (FUSS) which was theoretically
analysed in \cite{Hutter:01fuss} and then experimentally investigated
in \cite{Legg:04fussexp}.  FUSS works by focusing the selection
intensity on individuals which have uncommon fitness values rather
than on those with highest fitness as is usually done.  In this way a
broad range of individuals with many different levels of fitness
develop in the population including, hopefully, some individuals of
high fitness.  While FUSS achieved some interesting results, in
particular on highly deceptive problems, it also had difficulties in
some situations.  By focusing selection on rare individuals within the
population there was a tendency for these individuals to fill the
population with highly related offspring \iflonger (see
\cite{Legg:04fussexp} for a more complete explanation). \else
\cite{Legg:04fussexp}. \fi

\vspace{0.05em}

Here we take the idea of using fitness to roughly approximate the
similarity of individuals but apply it in a different way.  Rather
than using it to control selection for \emph{reproduction}, as FUSS
does, we instead use it to control selection for \emph{deletion},
hence the name Fitness Uniform Deletion Scheme or FUDS.  The result is
an easily implemented, computationally efficient and problem
independent approach to population diversity control.

\vspace{0.05em}

In \emph{Section \ref{sec:FUDS}} we describe the intuition behind FUDS
and its essential properties.  \emph{Section \ref{sec:Zeta}} details
the test system and how the tests were performed.  In \emph{Section
\ref{secEx}} we compare FUDS to random deletion on a deceptive 2D
optimisation problem.  \emph{Section \ref{sec:dtsp}} examines the
performance of FUDS and random deletion for random travelling salesman
problems with various selection intensities and population sizes.
\emph{Section \ref{sec:scp}} repeats this comparison for a set
covering problem.  \emph{Section \ref{sec:SAT}} looks at CNF3 SAT
problems and includes an analysis of population diversity using
hamming distance.  Finally \emph{Section \ref{sec:conc}} contains a
brief summary and possible future work.

\section{Fitness Uniform Deletion}\label{sec:FUDS}

The intuition behind FUDS is very simple: If an individual has a
fitness value which is very rare in the population then this
individual almost certainly contains unique information which, if it
were to be deleted, would decrease the total population diversity.
Conversely, if a large subset of individuals in the population all
have the same fitness then we may delete from this set without losing
much population diversity.  Presumably these individuals are common in
some sense and likely exist in parts of the solution space which are
easy to reach.

\vspace{0.2em}

From this simple observation FUDS immediately follows: \emph{Only
delete individuals which have very commonly occurring fitness.}  This
should help preserve population diversity, even for the most deceptive
problems.  Indeed it is now simply impossible for the whole population
to collapse to a collection of highly related individuals with similar
fitness.  The technique is simple to understand, easy to implement,
computationally efficient and completely independent of both the
problem and of the genotype representation being used.

\vspace{0.2em}

We implement FUDS as follows.  Let $f_{min}$ and $f_{max}$ be the
minimum and maximum fitness values possible for a problem, or at least
reasonable upper and lower bounds.  We divide the interval
$[f_{min},f_{max}]$ into a collection of subintervals of equal length
$\{ [f_{min},f_{min} + \varepsilon ), [f_{min} + \varepsilon ,f_{min}
+ 2\varepsilon ), \ldots, [f_{max}-\varepsilon, f_{max}] \}$.  We call
these intervals \emph{fitness levels}.  As individuals are added to
the population their fitness is computed and they are placed in the
list of individuals corresponding to the fitness level they belong to.
Thus the number of individuals in each fitness level describes how
common fitness values within this interval are in the current
population.  When a deletion is required the algorithm locates the
fitness level with the largest number of individuals and then randomly
selects an individual that belongs to this level for deletion.  In the
case of multiple fitness levels having the same size, the lowest
fitness level is chosen.

\vspace{0.2em}

If the number of fitness levels is chosen too low relative to the
population size, for example 5 fitness levels with a population of
500, then the resulting model of the distribution of individuals
across the fitness range will be too coarse.  Alternatively, if a
large number of fitness levels is used with a very small population
the individuals may become too thinly spread across the fitness
levels.  While in these extreme cases this could effect the
performance of FUDS, in practice we have found that the system is not
particularly sensitive to the setting of this parameter.  If $n$ is
the population size then setting the number of fitness levels to be
$\sqrt{n}$ is a good rule of thumb.  Generally, so long as the
population size is in the range of 2 to 50 times the number of fitness
levels performance is unaffected.  A theoretically better justified
solution would be to model the population distribution using
(in)finite Bayesian trees \cite{Hutter:04bayestree}.

\begin{figure}
\includegraphics[width=\columnwidth, height=0.22\textheight]{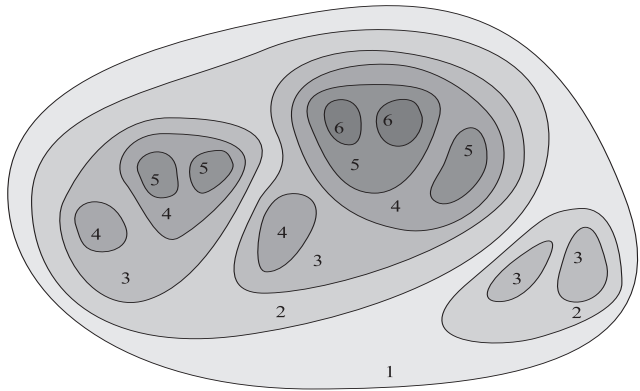}
\caption{\label{figtree} This illustrates the density of individuals
over a fitness landscape with FUDS.  The numbers represent the fitness
of the different regions and the shading represents the density of
individuals over the space.}
\end{figure}

\vspace{0.2em}

Under FUDS the takeover of the highest fitness level, or indeed any
fitness level, is impossible.  This is easy to see because as soon as
any fitness level starts to dominate, all of the deletions become
focused on this level until it is no longer the most populated fitness
level.  As a by-product, this also means that individuals on
relatively unpopulated fitness levels are preserved.
This allows the steady creation of individuals on many different
fitness levels and makes it relatively easy for the EA to find its way
out of local optima as it keeps on exploring evolutionary paths which
do not at first appear to be promising.

It is instructive to consider how individuals are distributed in the
genome space when using FUDS.  If we consider a distance semi-metric
between individuals $d(i,j) := | f(i) - f(j) |$ where $f$ is the
fitness function, then clearly FUDS attempts to uniformly distribute
the individuals according to this metric.  However this does not imply
that the individuals are uniformly distributed across the genome
space.  Typically, areas of high fitness are relatively small compared
to areas of lower fitness.  In this case, if we have the same number
of individuals in both regions then the density of individuals in the
high fitness region will be much higher due to its smaller size.  We
illustrate this visually on a fitness landscape in
Figure~\ref{figtree}.  The numbers indicate the fitness of individuals
in each region, while the density of individuals is given by how dark
the regions are shaded.

Clearly then, uniformity in the fitness distance metric $d$ does have
some implications for the distribution of individuals with respect to
a metric $g$ on the genome space.  This allows us to relate FUDS to
other methods of diversity control that require a genome space metric.
We say that a fitness function $f$ is \emph{smooth} with respect to
$g$, if $g( i, j )$ being small implies that $|f(i) - f(j)|$ is also
small, that is, $d( i, j )$ is small.  This implies that if $d( i, j
)$ is not small, $g( i, j )$ also cannot be small.  Thus, if we limit
the number of $d$ similar individuals, as we do in FUDS, this will
also limit the number of $g$ similar individuals, as is done in
crowding and niching methods.  The advantage of FUDS is that we do not
need to know what $g$ is, or to compute its value, something that can
be very difficult in some applications.  Indeed, the above argument is
true for \emph{any} metric $g$ on the genome space that $f$ is smooth
with respect to.  On the other hand, if the fitness function $f$ is
not smooth with respect to $g$, then this argument cannot be made.
However, in this case the optimisation problem is difficult as small
mutations in genome space with respect to $g$ will produce
unpredictable changes in fitness.

Because FUDS is only a deletion scheme, we still need to choose a
selection scheme which may require us to set a selection intensity
parameter.  With FUDS however, the performance of the system is less
sensitive to the correct setting of this parameter.  If the selection
intensity is set too high the normal problem is that the population
rushes into a \mbox{local} optima too soon and becomes stuck before it
has had a chance to properly explore the genotype space for other
promising regions.  However, with FUDS a total collapse in population
diversity is impossible and thus much higher levels of selection
intensity may be used.

Conversely, if the selection intensity is too low, the population
tends not to explore the higher areas of the fitness landscape at all.
Consider a population which contains 1,000 individuals.  Under random
deletion all of these individuals, including the highly fit ones, will
have a 1 in 1,000 chance of being deleted in each cycle and so the
expected life time of an individual is 1,000 deletion cycles.  Thus if
a highly fit individual is to contribute a child of the same fitness
or higher, it must do so reasonably quickly.  However for some
optimisation problems the probability of a fit individual having such
a child when it is selected is very low, so low in fact that it is
more likely to be deleted before this happens.  As a result the
population becomes stuck, unable to find individuals of greater
fitness before the fittest individuals are killed off.  With FUDS, as
rare fit individuals are not deleted, we can use much lower selection
intensity without the population becoming stuck.

\section{FUDS Test System}\label{sec:Zeta}

In order to test FUDS we have implemented a simple population based
optimiser in Java on a PC running Linux.  The full source code for
this along with usage instructions, example optimisation problems and
test data sets is available from\iffinal \cite{Legg:website}\else
[\emph{link removed for blind review}]\fi.  This zip file also
contains directions on where to download the full datasets that were
used to produce the results presented in this paper.

Our optimiser uses a ``steady state'' population rather than the more
usual ``generational'' population.  With a steady state population
individuals are selected and acted upon one at a time rather than in
bulk as under the generational approach.  Specifically the following
occurs: An individual is first selected according to some
\emph{selection scheme}, then according to the \emph{crossover
probability} a second individual may be selected and the two are
crossed to form a child.  Then according to the \emph{mutate
probability} a mutation operation may be applied to the child.  When a
crossover does not occur a mutation always takes place in order to
reduce the probability of a clone of an existing individual being
added to the population.  Finally the \emph{deletion scheme} selects
which individual from the population the new child will replace.

For the selection scheme we implemented the commonly used
\emph{tournament selection}.  Under tournament selection a group of
individuals is randomly chosen from the population, then the
individual with the highest fitness in this set is returned.  The size
of the group is called the \emph{tournament size} and it is clear that
the larger this group is the more likely we are to select a highly fit
individual from the population.  In our tests we have used tournament
sizes ranging from 2 to 12 in order to examine a wide range of
selection intensities.

We consider tournament selection to be roughly representative of other
standard selection schemes which favour the fitter individuals in the
population; indeed for tournament size 2 it can be shown that
tournament selection is equivalent to the linear ranking selection
scheme \cite[Sec.2.2.4]{Hutter:92cfs}. For other standard selection
schemes we expect the performance of these schemes to be at best
comparable to tournament selection when used with a correctly tuned
selection intensity.

Good values for the crossover and mutate probabilities depend on the
problem and must be manually tuned based on experience as there are
few theoretical guidelines on how to do this.  For some problems
performance can be quite sensitive to these values while in others
their values do not make much difference.  Our default values are 0.5
for both which has often provided us with reasonable performance.

With steady state optimisers the standard deletion scheme used is
simply random deletion.  The rational for this is that it is neutral
in the sense that it does not skew the distribution of the population
in any way.  Thus whether the population tends toward high or low
fitness etc.\ is solely a function of the selection scheme and its
parameters, in particular the selection intensity.  Of course random
deletion makes no effort to preserve diversity in the population as
all individuals have an equal chance of being removed.  Essentially
our objective in this paper is to investigate whether FUDS might be a
better alternative and under which circumstances.

When reporting test results we will adopt the following notation:
TOUR2 means tournament selection with a tournament size of 2.
Similarly for TOUR3, TOUR4 and so on.  When a graph shows the
performance of tournament selection over a range of tournament sizes
we will simply write TOURx.  To indicate the deletion scheme used, we
will add either the suffix -R or -F to indicate random deletion or
FUDS respectively.  Thus, TOUR10-R is tournament selection with a
tournament size of 10 used with random deletion.

For each problem we run the system using tournament selection with the
same tournament sizes, the same mutation and crossover rates and the
same population size.  The only difference is which deletion scheme is
used by the code.  Thus even if our parameters, mutation operators
etc.\ are not optimal for a given problem, the comparison between the
two deletion schemes is still fair.  Indeed we will often be
deliberately setting the optimisation parameters to non-optimal values
in order to compare the robustness of the system when using the
different deletion schemes.

As a steady state optimiser operates on just one individual at a time,
the number of cycles within a given run can be high, perhaps 100,000
or more.  In order to make our results more comparable to a
generational optimiser we divide this number by the size of the
population to give the approximate number of generations.
Unfortunately the theoretical understanding of the relationship
between steady state and generation optimisers is not strong.  It has
been shown that under the assumption of no crossover the effective
selection intensity using tournament selection with size 2 is
approximately twice as strong under a steady state GA as it is with a
generational GA \cite{Rogers:99}.  As far as we are aware a similar
comparison for systems with crossover has not been performed, though
we would not expect the results to be significantly different.

In each run of the system we stopped the optimiser after no progress
had been recorded for 20 generations.  Running the system longer and
looking at the graphs it seems that 20 generations is sufficient to
identify when the system becomes stuck and further progress is
unlikely.  In order to generate reliable statistics we ran each test
multiple times; typically 50 times.  From these runs we then
calculated the average performance for each selection scheme.  We also
computed the sample standard deviation and from this the standard
error in our estimate of the mean.  This value was then used to
generate 95\% confidence intervals which appear as error bars on the
graphs.

\section{A Deceptive 2D Problem}\label{secEx}

\begin{figure}
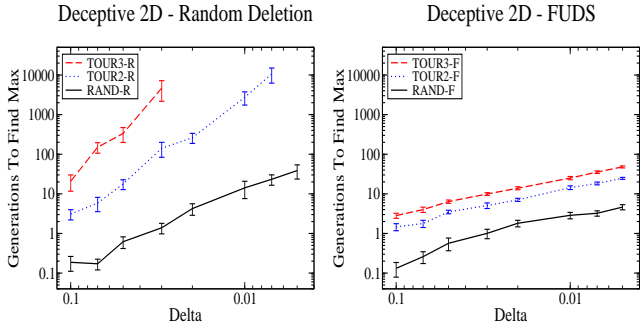

\includegraphics[width=0.232\textwidth,height=0.24\textwidth]{Deceptive-R-gecco.eps}
\hspace*{0.002\textwidth}
\includegraphics[width=0.232\textwidth,height=0.24\textwidth]{Deceptive-F-gecco.eps}
\caption{\label{SimpleProb} Switching from random deletion (left
graph) to FUDS (right graph) the number of generations required to
find the global optimum falls dramatically, especially when higher
selection intensity is used.}
\end{figure}

The first problem we examine is the highly deceptive 2D optimisation
problem previously analysed for FUSS in \cite{Hutter:01fuss}.  The
space of individuals is the unit square $[0,1]\times[0,1]$.  On this
space narrow regions $I_1 := [a,a+\delta] \times [0,1]$ and $I_2 :=
[0,1] \times [b,b+\delta]$ for some $a,b,\delta \in [0,1]$ are
defined.  Typically $\delta$ is chosen so that it is much smaller than
1 and thus $I_1$ and $I_2$ do not occupy much of the domain space.
The fitness function is defined by the equation,
$$
  f(x,y) = \left\{
  \begin{array}{l}
    1 \quad\mbox{if}\quad (x,y)\in I_1\backslash I_2, \\
    2 \quad\mbox{if}\quad (x,y)\in I_2\backslash I_1, \\
    3 \quad\mbox{if}\quad (x,y)\not\in I_1\cup I_2, \\
    4 \quad\mbox{if}\quad (x,y)\in I_1\cap I_2. \\
  \end{array}\right.
\parbox{3cm}
{\hfill \unitlength=0.6mm
\begin{picture}(45,45)
\scriptsize
\put(5,5){\vector(0,1){40}}
\put(5,5){\vector(1,0){40}}
\put(20,5){\line(0,1){35}}
\put(25,5){\line(0,1){35}}
\put(40,5){\line(0,1){35}}
\put(5,15){\line(1,0){35}}
\put(5,20){\line(1,0){35}}
\put(5,40){\line(1,0){35}}
\put(22.5,17.5){\makebox(0,0)[cc]{4}}
\put(12.5,30){\makebox(0,0)[cc]{3}}
\put(22.5,30){\makebox(0,0)[cc]{1}}
\put(32.5,30){\makebox(0,0)[cc]{3}}
\put(32.5,10){\makebox(0,0)[cc]{3}}
\put(12.5,10){\makebox(0,0)[cc]{3}}
\put(12.5,17.5){\makebox(0,0)[cc]{2}}
\put(32.5,17.5){\makebox(0,0)[cc]{2}}
\put(22.5,10){\makebox(0,0)[cc]{1}}
\put(44,2.5){\makebox(0,0)[cc]{$x$}}
\put(22.5,2.5){\makebox(0,0)[cc]{$\delta$}}
\put(20,3.5){\makebox(0,0)[cc]{$a$}}
\put(2.5,17.5){\makebox(0,0)[cc]{$\delta$}}
\put(4,14.5){\makebox(0,0)[cc]{$b$}}
\put(40,3){\makebox(0,0)[cc]{1}}
\put(3.5,40){\makebox(0,0)[cc]{1}}
\put(2.5,44){\makebox(0,0)[cc]{$y$}}
\put(22.5,42.5){\makebox(0,0)[cc]{$f(x,y)$}}
\end{picture}
}
$$

For this problem we set up the mutation operator to randomly replace
either the $x$ or $y$ position of an individual and the crossover to
take the $x$ position from one individual and the $y$ position from
another to produce an offspring.  The size of the domain for which the
function is maximised is just $\delta^2$ which is very small for small
values of $\delta$, while the local maxima at fitness level 3 covers
most of the space.  Clearly the only way to reach the global maximum
is by leaving this local maxima and exploring the space of individuals
with lower fitness values of 1 or 2.  Thus, with respect to the
mutation and crossover operators we have defined, this is a deceptive
optimisation problem as these partitions mislead the EA (see
\cite{Forrest:93} for a definition of ``deceptive'').

For this test we set the maximum population size to 1,000 and made 20
runs for each $\delta$ value.  With a steady state EA it is usual to
start with a full population of random individuals.  However for this
particular problem we reduced the initial population size down to just
10 in order to avoid the effect of doing a large random search when we
created the initial population and thereby distorting the scaling.
Usually this might create difficulties due to the poor genetic
diversity in the initial population.  However due to the fact that any
individual can mutate to any other in just two steps this is not a
problem in this situation.  Initial tests indicated that reducing the
crossover probability from 0.5 to 0.25 improved the performance
slightly and so we have used the latter value.

\begin{figure}
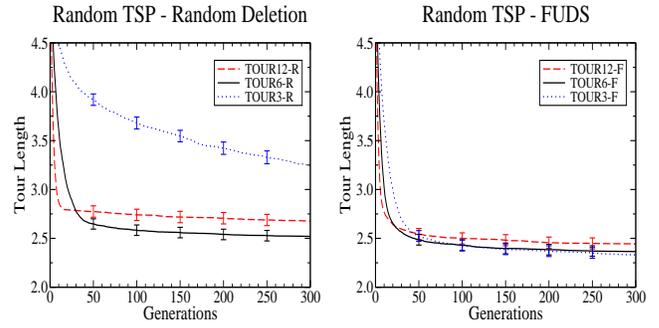

\includegraphics[width=0.232\textwidth,height=0.24\textwidth]{DTSPI-300gen-R-gecco.eps}
\hspace*{0.002\textwidth}
\includegraphics[width=0.232\textwidth,height=0.24\textwidth]{DTSPI-300gen-F-gecco.eps}
\caption{\label{DTSP-1} TOUR3-R converged too slowly while TOUR12-R
converged prematurely and became stuck.  TOUR6-R appears to be about
the correct tournament size for this problem.  With FUDS all of the
selection schemes performed well.}
\end{figure}

The first set of results for the selection schemes used with random
deletion appear in the left graph of Figure~\ref{SimpleProb}.  As
expected higher selection intensity is a significant disadvantage for
this problem.  Indeed even with just a tournament size of 3 the number
of generations required to find the maximum became infeasible to
compute for smaller values of $\delta$.  Be aware that this is a
log-log scaled graph and so the different slopes indicate
significantly different orders of scaling.  In the second set of tests
we switch from random deletion to FUDS.  These results appear in the
right graph of Figure~\ref{SimpleProb}.  We see that with FUDS as the
deletion scheme the scaling improves dramatically for RAND, TOUR2 and
TOUR3.

Although this problem was artificially constructed, the results
clearly demonstrate how for some very deceptive problems much higher
levels of selection intensity can be applied when using FUDS.

\section{Travelling Salesman Problem}\label{sec:dtsp}

A well known optimisation problem is the so called Travelling Salesman
Problem (TSP).  The task is to find the shortest Hamiltonian cycle
(path) in a graph of $N$ vertexes (cities) connected by edges of
certain lengths.  There exist highly specialised population based
optimisers which use advanced mutation and crossover operators and are
capable of finding paths less than one percent longer than the optimal
path for up to $10^7$ cities
\cite{Lin:73,Martin:96,Johnson:97,Applegate:00}.  As our goal is only
to study the relative performance of selection and deletion schemes,
having a highly refined implementation is not important.  Thus the
mutation and crossover operators we used were quite simple: Mutation
was achieved by just switching the position of two of the cities in
the solution, while for crossover we used the partial mapped crossover
technique~\cite{Goldberg:85}.  Fitness was computed by taking the
reciprocal of the tour length.

We have used randomly generated TSP problems, that is, the distance
between any two cities was chosen uniformly from the unit interval
$[0,1]$.  We chose this as it is known to be a particularly deceptive
form of the TSP problem as the usual triangle inequality relation does
not hold in general.  For example, the distance between cities $A$ and
$B$ might be $0.1$, between cities $B$ and $C$ $0.2$, and yet the
distance between $A$ and $C$ might be $0.8$.  The problem still has
some structure though as efficient partial solutions tend to be useful
building blocks for efficient complete tours.

\begin{figure*}[t]
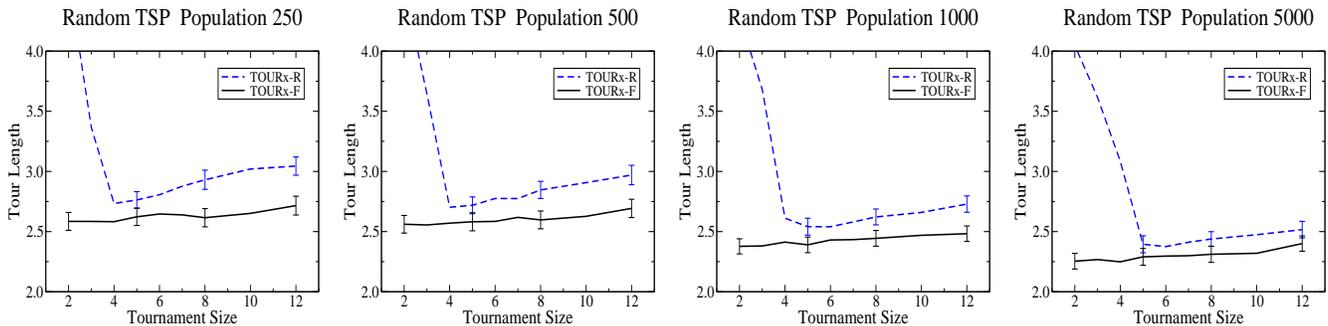

\includegraphics[width=0.2355\textwidth,height=0.24\textwidth]{dtpsi20-s40-p250-gecco.eps}
\hspace*{0.004\textwidth}
\includegraphics[width=0.2355\textwidth,height=0.24\textwidth]{dtpsi20-s40-p500-gecco.eps}
\hspace*{0.004\textwidth}
\includegraphics[width=0.2355\textwidth,height=0.24\textwidth]{dtpsi20-s40-p1k-gecco.eps}
\hspace*{0.004\textwidth}
\includegraphics[width=0.2355\textwidth,height=0.24\textwidth]{dtpsi20-s40-p5k-gecco.eps}
\caption{\label{tsp} In every situation FUDS was superior to random
deletion.  The performance with FUDS was also much more stable under
variation in the selection intensity.}
\end{figure*}

For this test we used random distance TSP problems with 20 cities and
a population size of 1000.  We found that changing the crossover and
mutation probabilities did not improve performance and so these have
been left at their default values of 0.5.  Our stopping criteria was
simply to let the GA run for 300 generations as this appeared to be
adequate for all of the methods to converge and allowed us to easily
graph performance versus generations in a consistent way for each
combination of selection and deletion scheme.

\vspace{0.15em}

The first graph in Figure~\ref{DTSP-1} shows each of the selection
schemes used with random deletion.  We see that TOUR3-R has
insufficient selection intensity for adequate convergence while
TOUR12-R quickly converges to a local optimum and then becomes stuck.
TOUR6-R has about the correct level of selection intensity for this
problem and population size.

\vspace{0.15em}

The second graph in Figure~\ref{DTSP-1} shows the same set of
selection schemes but now using FUDS as the deletion scheme.  With
FUDS the performance for all tournament sizes either stayed the same
or improved.  In the case of TOUR3 the improvement was dramatic and
for TOUR12 the improvement was also significant.  This is interesting
because it shows that with FUDS performance can improve when the
selection intensity is either too high or too low making the GA more
robust.  With TOUR3-R the selection intensity is low and thus we would
expect the population diversity to remain relatively strong.  Thus the
fact that TOUR3-F was so much better than TOUR3-R shows that FUDS can
have performance benefits due to not deleting rare fit individuals, as
was predicted earlier in Section~\ref{sec:FUDS}.

\vspace{0.15em}

\begin{figure*}[t]
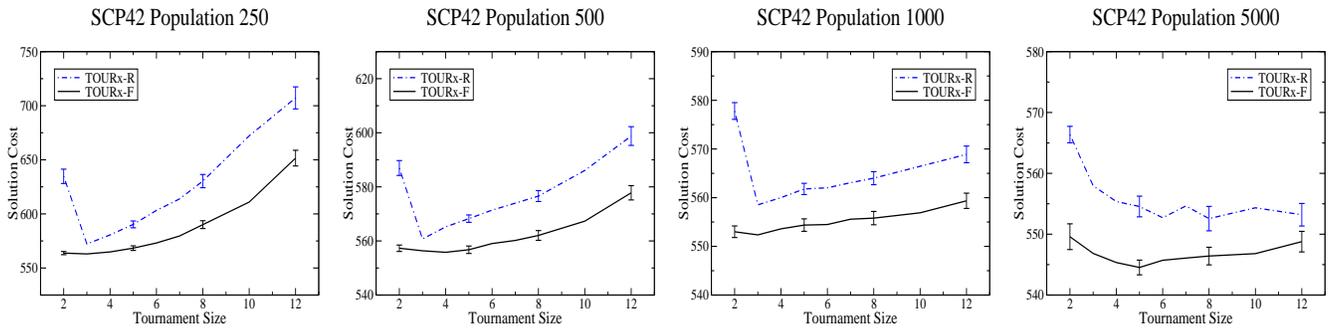

\includegraphics[width=0.2355\textwidth,height=0.24\textwidth]{SCPI42-p250-gecco.eps}
\hspace*{0.004\textwidth}
\includegraphics[width=0.2355\textwidth,height=0.24\textwidth]{SCPI42-p500-gecco.eps}
\hspace*{0.004\textwidth}
\includegraphics[width=0.2355\textwidth,height=0.24\textwidth]{SCPI42-p1k-gecco.eps}
\hspace*{0.004\textwidth}
\includegraphics[width=0.2355\textwidth,height=0.24\textwidth]{SCPI42-p5k-gecco.eps}
\caption{\label{scp-unbal} FUDS produced superior results to random
deletion in all situations tested, including when the tournament size
was optimally tuned.}
\end{figure*}

Investigating further it seems that this effect is due to the way that
FUDS focuses the deletion on the large mass of individuals which have
an average level of fitness while completely leaving the less common
fit individuals alone.  This helps a system with very weak selection
intensity move the mass of the population up through the fitness
space.  With higher selection intensity this problem tends not to
occur as individuals in this central mass are less likely to be
selected thus reducing the rate at which new individuals of average
fitness are added to the population.

\vspace{0.15em}

In order to better understand how stable FUDS performance is when used
with different selection intensities we ran another set of tests on
random TSP problems with 20 cities and graphed how performance varied
by tournament size.  For these tests we set the GA to stop each run
when no improvement had occurred in 40 generations.  We also tested on
a range of population sizes: 250, 500, 1000 and 5000.  The results
appear in Figure~\ref{tsp}.

In these graphs we can now clearly see how the performance of TOURx-R
varies significantly with tournament size.  Below the optimal
tournament size performance declined quickly while above this value it
also declined, though more slowly.  Interestingly, with a population
size of 5000 the optimal tournament size was about 6 while with small
populations this value fell to just 4.  Presumably this was partly
because smaller populations have lower diversity and thus cannot
withstand as much selection intensity.  In contrast, for every
combination of tournament size and population size the result with
FUDS was optimal.  \mbox{Indeed,} even with an optimally tuned
tournament size FUDS still improved performance.

More tests were run exploring performance with up to 100 cities.
Although the performance of FUDS remained much stronger than random
deletion for very low selection intensity, for high selection
intensity the two were equal.  We believe that the reason for this is
the following: When the space of potential solutions is very large,
finding anything close to a global optimum is practically impossible;
indeed it is difficult to even find the top of a reasonable local
optimum as the space has so many dimensions.  In these situations it
is more important to put effort into simply climbing in the space
rather than spreading out and trying to thoroughly explore.  Thus
higher selection intensity can be an advantage for large problem
spaces.  At any rate, for large problems and with high selection
intensity FUDS did not hinder the performance, while with low
selection intensity it continued to significantly improve it.

\section{Set Covering Problem}\label{sec:scp}

The set covering problem (\hspace{-0.05em}SCP\hspace{-0.05em}) is a
reasonably well known NP-complete optimisation problem with many real
world applications.  Let $M \in \{0,1\}^{m \times n}$ be a binary
valued matrix and let $c_j > 0$ for $j \in \{1, \ldots n \}$ be the
cost of column $j$.  The goal is to find a subset of the columns such
that the cost is minimised.  Define $x_j = 1$ if column $j$ is in our
solution and 0 otherwise.  The solution cost is then $\sum_{j=1}^n c_j
x_j$ subject to the condition that $\sum_{j=1}^n m_{ij} x_j \geq 1$
for $i \in \{1, \ldots m\}$.

Our system of representation, mutation operators and crossover follow
that used by Beasley \cite{Beasley:96} and we compute the fitness by
taking the reciprocal of the cost.  The results presented here are
based on the ``scp42'' problem from a standard collection of SCP
problems \cite{Beasley:03}.  The results obtained on other problems in
this test set were similar.  We found that increasing the crossover
probability and reducing the mutation probability improved
performance, especially when the selection intensity was low.  Thus we
have tested the system with a crossover probability of 0.8 and a
mutation probability of 0.2.  We performed each test at least 50 times
in order to minimise the error bars.  Our stopping criteria was to
terminate each run after no progress had occurred for 40 generations.
The results for this test appear in Figure~\ref{scp-unbal}.

Similar to the TSP graphs we again see the importance of correctly
tuning the tournament size with TOURx-R.  We also see the optimal
range of performance for TOURx-R moving to the right as the population
sizes increases.  This is what we would expect due to the greater
diversity in larger populations being able to support more selection
intensity.  This kind of variability is one of the reasons why the
selection intensity parameter usually has to be determined by
experimentation.

With FUDS the results were again very impressive.  As with the TSP
tests; for all combinations of tournament size and population size
that we tested, the performance with FUDS was superior to the
corresponding performance with random deletion.  This was true even
when the tournament size was not set optimally.  While the performance
of TOURx-F did vary with different tournament sizes, the results were
more robust than TOURx-R, especially with larger populations.  Indeed
for the larger two populations we again have a situation where the
worst performance of TOURx-F is equal to the best performance of
\mbox{TOURx-R}.

\section{Maximum CNF3 SAT}\label{sec:SAT}

Maximum CNF3 SAT is a well known NP hard optimisation problem
\cite{Crescenzi:04} that has been extensively studied.  A three
literal conjunctive normal form (CNF) logical equation is a boolean
equation that consists of a conjunction of clauses where each clause
contains a disjunction of three literals.  So for example, $(a \lor b
\lor \lnot c) \land ( a \lor \lnot e \lor f)$ is a CNF3 expression.
The goal in the maximum CNF3 SAT problem is to find an instantiation
of the variables such that the maximum number of clauses evaluate to
true.  Thus for the above equation if $a = F$, $b = T$, $c = T$, $e =
T$, and $f = F$ then just one clause evaluates to true and thus this
instantiation gets a score of one.  Achieving significant results in
this area would be difficult and this is not our aim; we are simply
using this problem as a test to compare deletion schemes.

\begin{figure}
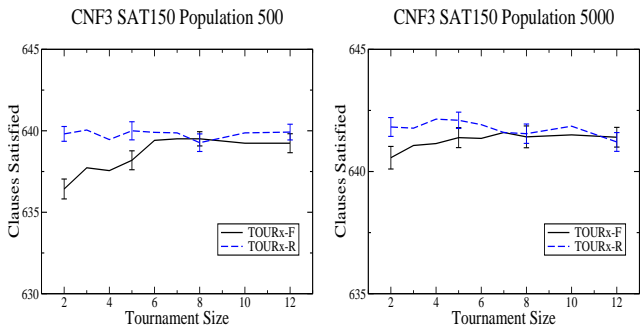

\centering
\includegraphics[width=0.231\textwidth,height=0.24\textwidth]{CNF150-p500-gecco.eps}
\hspace*{0.002\textwidth}
\includegraphics[width=0.231\textwidth,height=0.24\textwidth]{CNF150-p5k-gecco.eps}
\caption{\label{cnf-all}With low selection intensity TOURx-F performed
slightly below TOURx-R, but was otherwise comparable.}
\end{figure}

Our test problems have been taken from the SATLIB collection of SAT
benchmark tests \cite{Hoos:00}.  The first test was performed on the
full set of 100 instances of randomly generated CNF3 formula with 150
variables and 645 clauses, all of which are known to be satisfiable.
Based on test results the crossover and mutation probabilities were
left at the default values.  Our mutation operator simply flips one
boolean variable and the crossover operator forms a new individual by
randomly selecting for each variable which parent's state to take.
Fitness was simply taken to be the number of clauses satisfied.  As in
previous sections we tested across a range of tournament sizes and
population sizes.  The results of these tests appear in
Figure~\ref{cnf-all}.

\vspace{0.1em}

We have shown only the population sizes of 500 and 5,000 as the other
population sizes tested followed the same pattern.  Interestingly, for
this problem there was no evidence of better performance with FUDS at
higher selection intensities.  Nor for that matter was there the
decline in performance with TOURx-R that we have seen elsewhere.
Indeed with random deletion the selection intensity appeared to have
no impact on performance at all.  While SAT3 CNF is an NP hard
optimisation problem, this lack of dependence of our selection
intensity parameter suggests that it may not have the deceptive
structure that FUDS was designed for.

\vspace{0.12em}

With low selection intensity FUDS caused performance to fall below
that of random deletion; something that we have not seen before.
Because the advantages of FUDS have been more apparent with low
populations in other test problems, we also tested the system with a
population size of only 150. Unfortunately no interesting changes in
behaviour were observed.

\vspace{0.12em}

\begin{figure}
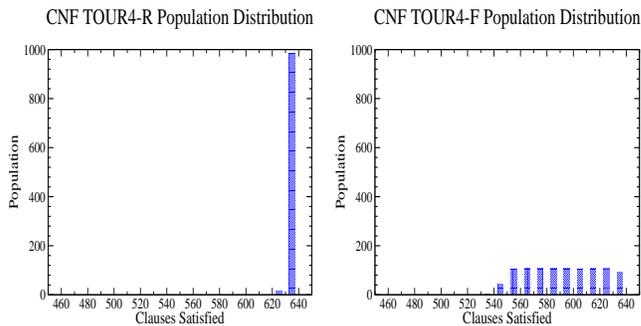

\centering
\includegraphics[width=0.231\textwidth,height=0.24\textwidth]{CNF-TOUR4-R-popDist.eps}
\hspace*{0.002\textwidth}
\includegraphics[width=0.231\textwidth,height=0.24\textwidth]{CNF-TOUR4-B-popDist.eps}
\caption{\label{cnf-pop} With TOUR4-R the population collapses to a
narrow band of fitness levels while with TOUR4-F the distribution is
flat.}
\end{figure}

We suspected that the uniform nature of the population distribution
that should occur with FUDS might be to blame as we only expect this
to be a benefit for very deceptive problems which are sensitive to the
tuning of the selection intensity parameter.  Thus we ran the EA with
a population of 1000 and graphed the population distribution across
the number of clauses satisfied at the end of the run.  We stopped
each run when the EA made no progress in 40 generations.  The results
of this appear in Figure~\ref{cnf-pop}.

\vspace{0.12em}

The first thing to note is that with TOUR4-R the population collapses
to a narrow band of fitness levels, as expected.  With TOUR4-F the
distribution is now uniform, though practically none of the population
satisfies fewer than 550 clauses.  The reason for this is quite
simple: While FUDS levels the population distribution out, TOUR4 tends
to select the most fit individuals and thus pushes the population to
the right from its starting point.

\vspace{0.12em}

Given that our goal is to find an instantiation that satisfies all 645
clauses, it is questionable whether having a large percentage of the
population unable to satisfy even 600 clauses is of much benefit.
While the total population diversity under FUDS might be very high,
perhaps the kind of diversity that matters the most is the diversity
among the relatively fit individuals in the population.  This should
be true for all but the most deceptive problems.  By thinly spreading
the population across a very wide range of fitness levels we actually
end up with very few individuals with the kind of diversity that
matters.  Of course this depends on the nature of the problem we are
trying to solve and the fitness function that we use.

With CNF3 SAT problems we can directly measure population diversity by
taking the average hamming distance between individuals' genomes.
While this means that the value of the fitness based similarity metric
is questionable for this problem, as more direct methods can be
applied, it is a useful situation for our analysis as it allows us to
directly measure how effective FUDS is at preserving population
diversity.  The hope of course is that any positive benefits that we
have seen here will also carry over to problems where directly
measuring the diversity is much more difficult.

\begin{figure}
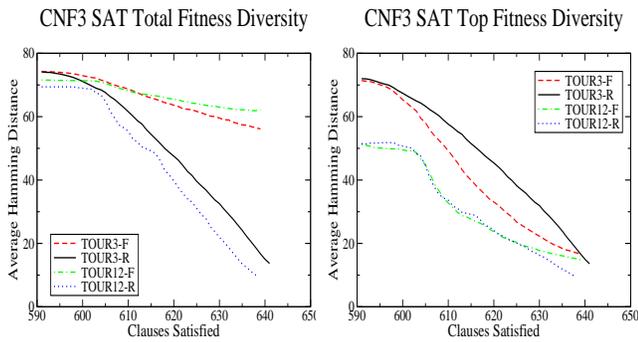

\includegraphics[width=0.236\textwidth,height=0.25\textwidth]{CNF-total-diversity-gecco.eps}
\includegraphics[width=0.236\textwidth,height=0.25\textwidth]{CNF-top-diversity-gecco.eps}
\caption{\label{cnf-diver} While the total population diversity is
improved by FUDS, the diversity among fit individuals is similar to
that with random deletion.}
\end{figure}

For the diversity tests we used a population size of 1000 again.  For
comparison we used TOUR3 and TOUR12 both with random deletion and with
FUDS.  In each run we calculated two different statistics: The average
hamming distance between individuals in the whole population, and the
average hamming distance between individuals whose fitness was no more
than 20 below the fittest individual in the population at the time.
These two measurements give us the ``total population diversity'' and
``top fitness diversity'' graphs in Figure~\ref{cnf-diver}.

We graphed these measurements against the number of clauses satisfied
by the fittest individual rather than the number of generations.  This
is only fair because if good solutions are found very quickly then an
equally rapid decline in diversity is acceptable and to be expected.
Indeed it is trivial to come up with a system which always maintains
high population diversity however long it runs, but is unlikely to
find any good solutions.  The results were averaged over all 100
problems in the test set.  Because the best solution found in each run
varied, we have only graphed each curve until the point where fewer
than 50\% of the runs were able to achieve this level of fitness.
Thus the terminal point at the right of each curve is representative
of fairly typical runs rather than just a few exceptional ones that
perhaps found unusually good solutions by chance.

\mbox{}
\vspace{-0.6em}

The left graph in Figure~\ref{cnf-diver} shows the total population
diversity.  As expected the diversity with TOUR3-R and TOUR12-R
decline steadily as finding better solutions becomes increasingly
difficult and the population tends to collapse into a narrow band of
fitness.  Also the total population diversity with TOUR3-R is higher
than with TOUR12-R as we would expect.  Importantly, FUDS
significantly improved the total population diversity with both TOUR3
and TOUR12 as desired.  However because the maximal solution found by
TOUR3-F and TOUR12-F were not better than TOUR3-R and TOUR12-R this
indicates that improved total population diversity was not a
significant factor for this optimisation problem.

On the right graph we see the diversity among the fitter individuals
in the population.  TOUR3 has significantly greater diversity than
TOUR12 with both deletion schemes.  This is expected as TOUR3 tends to
search more evolutionary paths while TOUR12 just rushes down a few.
Disappointingly FUDS does not appear to have made very much difference
to the diversity among these highly fit individuals, though the curves
do appear to flatten out a little as the diversity drops below 30, so
perhaps FUDS is having a slight impact.

In summary, these results show that while FUDS has been successful in
maximising total population diversity, for problems such as CNF3 SAT
this in itself is not sufficient.  It appears to be more important
that the GA maximises the diversity among those individuals which have
reasonably high fitness.

\section{Conclusions and Future Work}\label{sec:conc}

We have used tournament selection to test FUDS against random deletion
on several optimisation problems with different population sizes,
mutation probabilities and crossover probabilities.  For the
artificial deceptive 2D problem, random distance matrix TSP problems
and the SCP problem, FUDS was consistently superior, returning better
results than random deletion for every combination of tournament size
and population size tested.  This is particularly significant given
that FUDS is trivial to implement, computational cheap and largely
problem independent.

For CNF3 SAT problems however the results were less impressive.  While
the performance with FUDS was comparable to random deletion for medium
to high selection intensity, it was inferior to random deletion for
low selection intensities.  Investigating further we found that while
the total population diversity was improved, as expected, the
diversity among the fit individuals was not.  However, because the
performance of TOURx-R was not reduced with very high selection
intensity, this indicates that CNF3 SAT problems do not have the kind
of diversity problems that FUDS was designed to overcome.  That is,
problems with serious population diversity difficulties where greedy
exploration is harshly punished.

In future work FUDS should be tested on more problem classes with an
aim to developing a better understanding of what kinds of deceptive
optimisation problems it is the most effective.\\

\subsection*{Acknowledgements}

This work was supported by SNF grant 2100-67712.02.\\

\end{document}